\documentclass[letterpaper]{article} 
\usepackage{aaai25}  
\usepackage{times}  
\usepackage{helvet}  
\usepackage{courier}  
\usepackage[hyphens]{url}  
\usepackage{graphicx} 
\usepackage{natbib}  
\usepackage{caption} 
\usepackage{algorithm}
\usepackage{algorithmic}
\usepackage{booktabs}
\usepackage{pifont}
\usepackage{multirow}
\usepackage{amsmath}

\usepackage{newfloat}
\usepackage{listings}
\DeclareCaptionStyle{ruled}{labelfont=normalfont,labelsep=colon,strut=off} 
\floatstyle{ruled}
\newfloat{listing}{tb}{lst}{}
\floatname{listing}{Listing}
\title{VE-Bench: Subjective-Aligned Benchmark Suite for Text-Driven Video Editing Quality Assessment}

\author{
    Shangkun Sun\textsuperscript{\rm 1, \rm 2},
    Xiaoyu Liang\textsuperscript{\rm 1},
    Songlin Fan\textsuperscript{\rm 1,2},
    Wenxu Gao\textsuperscript{\rm 1,2},
    Wei Gao\textsuperscript{\rm 1,2}\thanks{Corresponding author.}
}
\affiliations{
    \textsuperscript{\rm 1}SECE, Peking University,
    \textsuperscript{\rm 2}Peng Cheng Laboratory \\
    \{sunshk, liangxiaoyu\_yy, gaowx\}@stu.pku.edu.cn, \{slfan, gaowei262\}@pku.edu.cn
}

\begin{document}
\maketitle

\begin{abstract}
Text-driven video editing has recently experienced rapid development. Despite this, evaluating edited videos remains a considerable challenge. Current metrics tend to fail to align with human perceptions, and effective quantitative metrics for video editing are still notably absent. To address this, we introduce VE-Bench, a benchmark suite tailored to the assessment of text-driven video editing. 
This suite includes VE-Bench DB, a video quality assessment (VQA) database for video editing. VE-Bench DB encompasses a diverse set of source videos featuring various motions and subjects, along with multiple distinct editing prompts, editing results from 8 different models, and the corresponding Mean Opinion Scores (MOS) from 24 human annotators. Based on VE-Bench DB, we further propose VE-Bench QA, a quantitative human-aligned measurement for the text-driven video editing task. 
In addition to the aesthetic, distortion, and other visual quality indicators that traditional VQA methods emphasize, VE-Bench QA focuses on the text-video alignment and the relevance modeling between source and edited videos. 
It proposes a new assessment network for video editing that attains superior performance in alignment with human preferences.
To the best of our knowledge, VE-Bench introduces the first quality assessment dataset for video editing and an effective subjective-aligned quantitative metric for this domain. 
All data and code will be publicly available to the community.
\end{abstract}

\section{Introduction}

\begin{figure}[t]
\centering
\includegraphics[width=1.0\columnwidth]{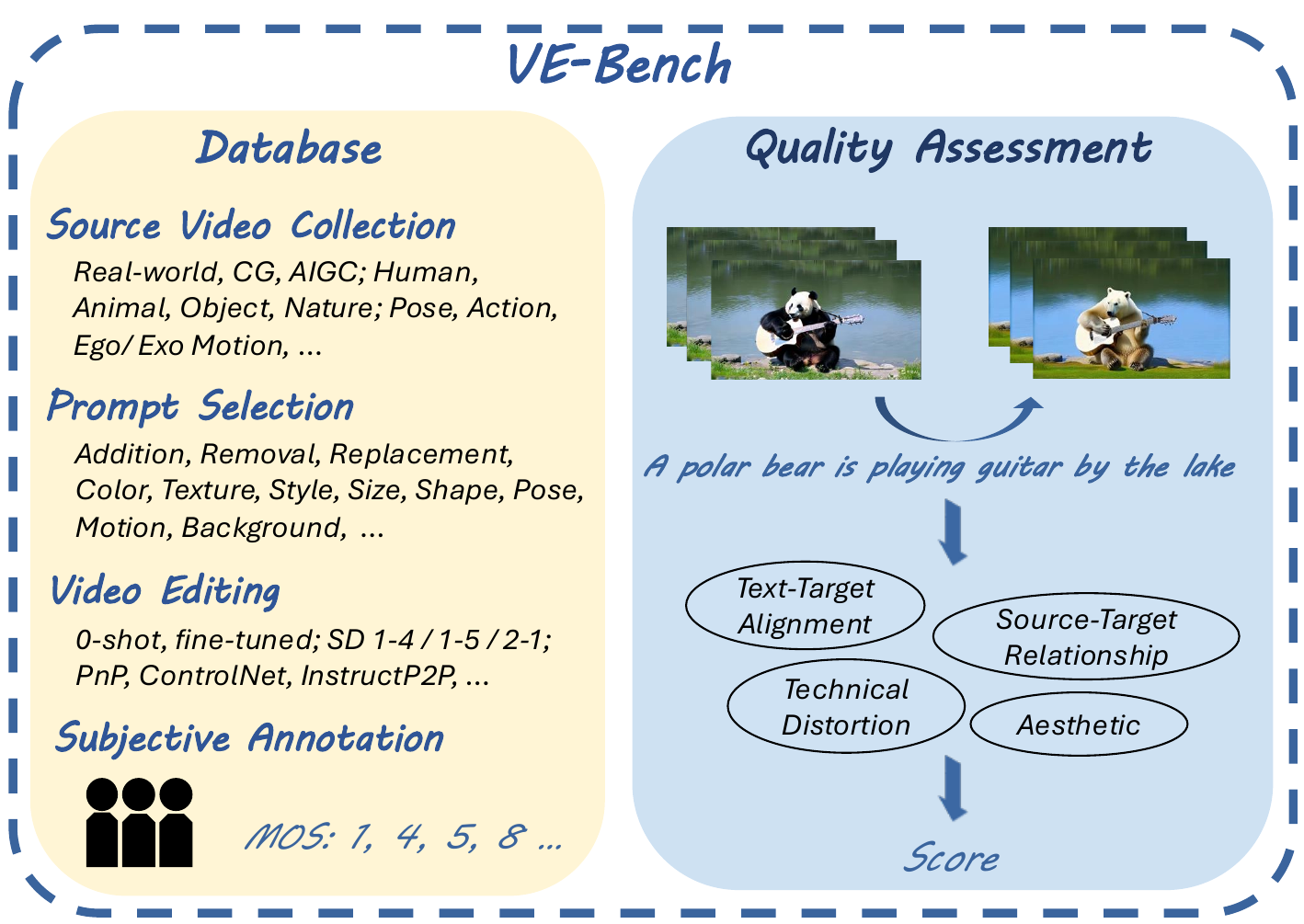} 
\caption{Overview of the proposed VE-Bench.}
\label{fig:1}
\end{figure}

With the rise of the AI-Generated Content (AIGC) trend, an increasing number of text-driven video editing methods~\cite{ma2024followyourpose,ma2024followyourclick,ma2023magicstick,ma2024followyouremoji,feng2024dit4edit} are gaining momentum and finding widespread application in daily life. However, there remains a lack of a suitable quantitative metric to assess video editing quality. Currently, the predominant evaluation method involves subjective experiments with human participants, which are costly and yield non-reusable results.

Currently, quantitative metrics such as CLIP and LPIPS scores~\cite{clip, lpips}, FVD~\cite{fvd}, and Warp scores~\cite{pix2video, rave} are commonly used in recent works. They mainly quantify results based on objective measurements in dimensions like editing quality, temporal consistency, and text consistency. Nevertheless, these existing metrics primarily face two main issues: \emph{(1) The tendency of misalignment with human subjective perceptions. (2) Incomplete evaluation. As they mainly measure results from a single dimension, making it difficult to comprehensively reflect the overall quality of the effects.}

Traditional video quality assessment (VQA) methods~\cite{dover, fastvqa, stablevqa} have been able to align with human perceptions. However, these methods are primarily designed for natural videos and struggle to evaluate AIGC video editing tasks adequately. In this case, these methods overlook the different distortions in AIGC videos~\cite{t2vqa}, such as irrational objects and irregular motion patterns. Besides, they do not simultaneously consider text-video alignment and the inner connection between source and edited videos, which are crucial for the assessment of editing results.

To address these challenges, we propose VE-Bench, a specialized suite tailored for text-driven AIGC video editing, as shown in Figure~\ref{fig:1}. 
We first establish VE-Bench DB, a quality assessment database for video editing. VE-Bench DB collects source videos containing real-world videos, CG-rendered videos, and AIGC videos, with multiple different actions, subjects, and scenarios, along with various types of prompts and edited results from different video editing methods. We then assembled 24 human subjects to gather the Mean Opinion Scores (MOS) for each video. To the best of our knowledge, this is the first AIGC VQA dataset for evaluating the quality of edited videos.

Building on this foundation, we introduce VE-Bench QA, a novel multi-modal quality assessment network specifically designed for AIGC video editing. VE-Bench QA evaluates edited videos from various aspects such as source-target video relationship, text-video alignment, and other aspects such as aesthetics and distortion. Detailed experiments demonstrate that VE-Bench QA achieves state-of-the-art alignment with human preferences, surpassing existing advanced metrics and VQA methods.

Our contributions could be summarized as: (1) We collect VE-Bench DB, a diverse dataset with videos featuring various motions and subjects, along with multiple distinct editing prompts, and the corresponding editing results with rich human feedback. To the best of our knowledge, it is the first quality assessment dataset for text-driven video editing. (2) Unlike traditional VQA methods that focus mainly on visual quality indicators, we propose VE-Bench QA, which further emphasizes text-video alignment and relevance modeling between source and edited videos. (3) The proposed VE-Bench QA is tailored to the assessment of the text-driven video editing task, surpassing existing advanced evaluation methods in aligning with human subjective ratings and showcasing the effectiveness of VE-Bench QA in evaluating AIGC video editing quality.

\section{Related Work}
\subsection{Metrics for Video Editing}
Currently, metrics commonly used in text-driven video editing include some objective metrics~\cite{clip,lpips,rave,fvd}, as well as some Video Quality Assessment (VQA) methods~\cite{hartwig2024evaluating,qalign,qbench,trivqa, wu2023human, imagereward} aligned with human feedback.
CLIP~\cite{clip} has been widely used due to its success in vision-language tasks~\cite{wu2024building, jia2024contexthoispatialcontextlearning, jia2024orchestratingsymphonypromptdistribution}.
CLIP-T calculates the average cosine similarity between each edited frame and the corresponding textual prompt.  
CLIP-F(Tmp-Con), refers to the average cosine similarity between consecutive edited frames.
Fram-Acc~\cite{fatezero} represents the percentage of edited frames that has a higher similarity to the target text than to the original source text.
LPIPS-P~\cite{stablevideo} and LPIPS-T~\cite{stablevideo} denote the LPIPS deviation from the original video frames and deviation between adjacent edited frames, respectively.
FVD~\cite{fvd} calculates the Fréchet Distance between two videos.
Spat-Con~\cite{fresco} refers to the average distance between VGG features. 
OSV (Objective Semantic Variance), proposed by~\cite{videop2p}, uses DINO-ViT~\cite{dino} to measure semantic consistency and calculates the frame-wise feature variance in the edited region. 
Warp-MSE and Warp-SSIM~\cite{rave} represent the MSE and SSIM between the edited video and the warped edited video by optical flow models. 
However, these individual metrics often only assess the editing results from a single dimension. 
$Q_{edit}$~\cite{rave} multiplies Warp-SSIM with CLIP-T to obtain a more comprehensive evaluation.
$S_{edit}$~\cite{flatten} (CLIP-T / Warp-MSE) combines Warp-MSE with CLIP-T to provide overall the assessment for videos. 
Nevertheless, these metrics are not aligned with human perceptions.
PickScore~\cite{pickscore} calculates the estimated alignment with human preferences via a CLIP-style model fine-tuned on human preference data.
FastVQA~\cite{fastvqa} proposes grid mini-patch sampling to evaluate videos efficiently via the consideration of local quality.
SimpleVQA~\cite{simplevqa} leverages quality-aware spatial features and motion features to assess videos. 
DOVER~\cite{dover} evaluates natural videos from the aesthetic and technical distortion perspective. 
However, these methods are typically suitable for evaluating single videos, neglecting the inherent relationship between edited results and the source video, and many VQA methods do not model the alignment between text and video. Currently, there is still a lack of a proper metric to evaluate the edited results based on the source video and editing prompts.

\subsection{Datasets for Video Editing Assessment}
In assessing edited videos, a common practice in prior works~\cite{fatezero,tune-a-video,fresco,sun2024t2v} has been to assemble human annotators to conduct subjective preference experiments. However, the results of subjective preference experiments are difficult to reproduce, and there is considerable variance in the data and prompts selected when comparing different methods. Recently, some studies~\cite{tvge, ccedit} have curated high-quality video-editing prompt pairs through diverse data collection and prompt design for unified community assessment. Nevertheless, these efforts still face two challenges:
(1) These datasets do not include subjective experimental feedback (Mean Opinion Score, MOS) corresponding to the video data, requiring others to still use objective metrics or conduct subjective experiments again when utilizing these datasets.
(2) The scenarios covered by these datasets could potentially be expanded further. For instance, TGVE~\cite{tvge} has only collected 76 videos, and BalanceCC~\cite{ccedit} has gathered 100 videos, both of which mainly focus on real-world scenes. In this work, we introduce VE-Bench DB, a dataset for assessing video editing quality that includes diverse content and human feedback scores. This dataset encompasses a variety of categories including real-world scenes, CG-rendered scenes, and AIGC-generated scenes. It covers various subjects such as people performing different actions, occupations, genders, and ages, as well as different animals, objects, and landscapes. The dataset also includes multiple types of motion, such as ego-motion and exo-motion, along with various editing prompts. This provides a solid foundation for more robust video editing evaluation. Details on the dataset are provided in the following sections. In total, VE-Bench DB comprises 169 different videos edited using 8 different video editing methods, yielding 1,170 edited results after manual screening. After that, 24 human subjects are invited to obtain the corresponding MOS scores. To the best of our knowledge, it is the first VQA dataset for text-driven video editing.

\subsection{Methods for Video Editing}
Recently, with the fast development of diffusion models~\cite{md3}, lots of video editing methods have emerged~\cite{tune-a-video, stablevideo,ma2022visual,chen2024follow,zhu2024instantswap, wang2024cove}. 
Different from image editing, one key to video editing is to maintain temporal consistency.
Tune-a-video~\cite{tune-a-video} inflates the 2D convolutions in T2I models to pseudo-3D convolutions and fine-tunes the attention matrix with source videos. Text2Video-Zero~\cite{t2v-zero} leverages cross-frame attention and introduces latent motion dynamics to keep the global scene and the background consistent. FateZero~\cite{fatezero} proposes to fuse the attention maps in the inversion
process and generation process and utilizes the source prompt’s cross-attention map to improve consistency. ControlVideo~\cite{controlvideo} introduces the interleaved-frame smoother and full-frame attention to keep the temporal consistency and model the temporal relationship in different frames.
Flatten~\cite{flatten} utilizes optical-flow-guided attention during the diffusion process to improve the visual consistency, which models the similarity of patch trajectories in different clips.
RAVE~\cite{rave} leverages the full-frame attention and proposes the grid sampling strategy to maintain the temporal consistency in video sequences. Based on Rerender-a-video~\cite{rerender-a-video} and the flow-guided attention in~\cite{flatten}, Fresco~\cite{fresco} further develops a group of temporal attention including efficient cross-frame attention, spatial/temporal-guided attention, etc., and fine-tunes the translated feature for better temporal consistency. In this work, we selected a variety of methods for video editing, including early and recent approaches, Zero-shot and few-shot techniques, different Stable Diffusion (SD) base models, and various editing methodologies, to ensure the diversity of the generated results.

\begin{figure*}[t]
\centering
\includegraphics[width=2.0\columnwidth]{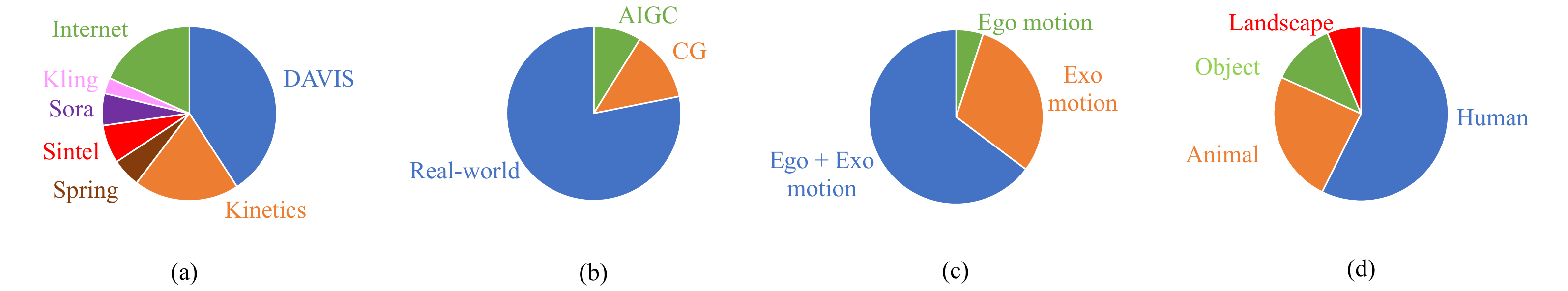} 
\caption{Collection of source videos. (a) Sources of videos. (b) Types of videos. (c) Motion categories. (d) Content categories.} 
\label{fig:src_video_collection}
\end{figure*}

\section{VE-Bench DB: Subjective-Aligned Dataset for Text-Driven Video Editing}
The collection of VE-Bench DB involves four primary stages: source video collection, prompt composition, selection and execution of video editing methods, and subjective experiments. We will discuss each part in subsequent sections.

\subsection{Source Video Collection}

To support a more robust quality assessment for video editing, VE-Bench collected a diverse set of source videos that are not limited to real-world scenes but also include some content rendered by computer graphics and text-driven AIGC videos. Notably, considering the current wide range of use cases, real-world scenes still account for a larger proportion. Different from previous works, given concerns about copyright issues, watermarks, and resolution, we did not randomly sample videos from Webvid~\cite{webvid}. Instead, to cover as many different content subjects, action categories, and scenarios as possible, VE-Bench manually selected 123 videos from four datasets: DAVIS~\cite{davis}, Kinetics-700~\cite{kinetics}, Sintel~\cite{sintel}, and Spring~\cite{spring}. Spring and Sintel are high-resolution datasets rendered by computer graphics. To ensure a diversity of actions and rich content, we did not randomly sample but carefully handpicked the corresponding video content. 
We tag each sample as Nature/Object/Animal/Human, Ego/Exo, etc., skipping similar or short videos until datasets are exhausted. We start from small datasets to large ones, and finally supplement from the Internet, which is vital to finding cases like auroras, lava, and lightning, as well as diverse action cases absent in traditional datasets.
In addition, we selected 15 different videos from Sora~\cite{sora} and Kling~\cite{kling} based on the principle of diversity in motion and content. Furthermore, to cover more content and a variety of actions, we selected 31 videos from the internet with the appropriate permissions. 
Ultimately, we collected 169 source videos with diverse content, and their specific sources, contents, and category compositions are shown in Figure~\ref{fig:src_video_collection}. All selected videos were resized to have a long side of 768 pixels while maintaining their original aspect ratios. Considering the limited length supported by existing video editing methods, each video was trimmed to 32 frames.

\subsection{Prompt Selection}
\begin{figure}[htbp]
\centering
\includegraphics[width=1.0\columnwidth]{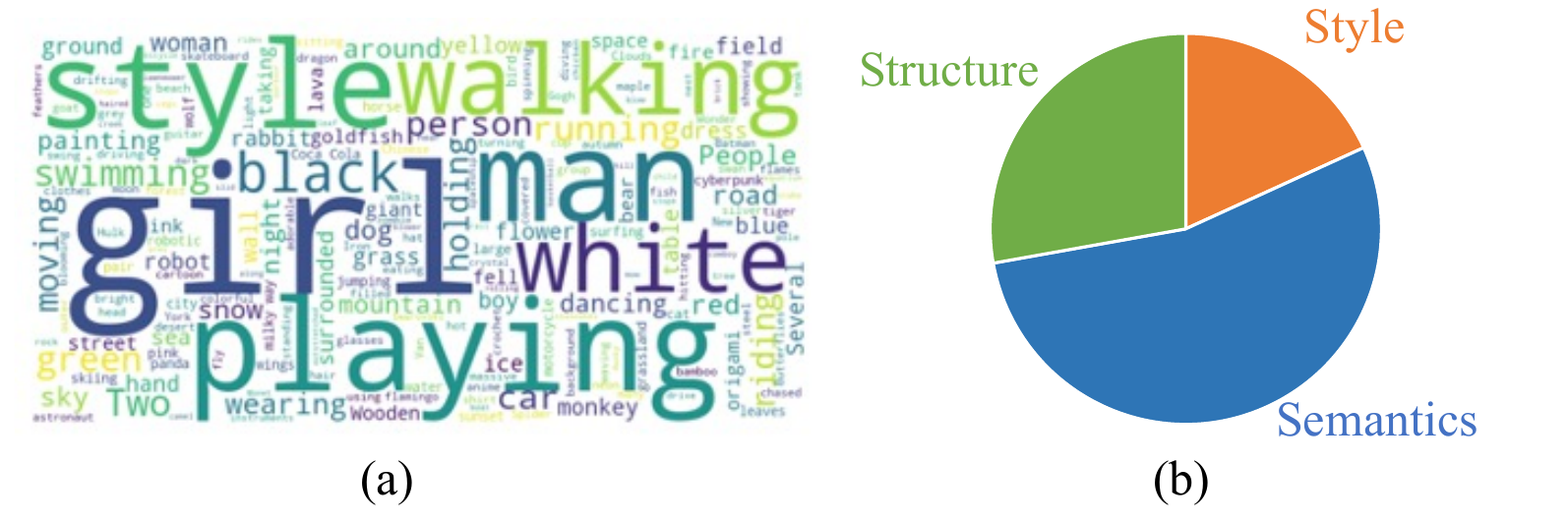} 
\caption{Statistics of VE-Bench DB prompts. (a) Word cloud of VE-Bench DB prompts. (b) Proportion of different types}

\label{fig:prompt_info}
\end{figure}

Referring to past work~\cite{prompt}, we classify prompts used for video editing into three major categories: (1) Style editing, which includes the edit on color, texture, or the overall atmosphere. (2) Semantic editing, which includes background editing and local editing such as the addition, replacement, or removal on a certain object. (3) Structural editing, which includes the change in object size, pose, motion, etc. To ensure the specificity and diversity of the prompts, we manually crafted corresponding prompts for each video, and the specific distribution is shown in Figure~\ref{fig:prompt_info}.

\subsection{Video Editing}

We then select 8 video editing methods. To ensure the distribution of edited video quality, in addition to recent top-performance models, we also include some earlier video editing methods. Besides, we select methods with different base models ranging from SD 1-4 to SD2-1 to improve the diversity of edited results.
Furthermore, to ensure the diversity of edited content, we choose both Zero-shot methods and methods that require fine-tuning. We also select models based on different editing paradigms, including effective editing strategies such as Instruct P2P~\cite{instructpix2pix}, PnP~\cite{pnp}, ControlNet~\cite{controlnet}, etc.
The specific details are presented in Table~\ref{tab:editing_methods}.

\begin{table*}[htb]
\centering
\scalebox{0.9}{
\begin{tabular}{llccc}
\toprule
Model & Time & Zero-shot & Edit. & SD. \\ \midrule
Tune-a-video~\cite{tune-a-video}      &  ICCV'23   &    \ding{55}   &   Others  &    1-4       \\ \midrule
T2V-Zero~\cite{t2v-zero}      &  ICCV'23    &    \ding{51}       &     Instruct-P2P~\cite{instructpix2pix}       &   1-5         \\ \midrule
Fate-Zero~\cite{fatezero}      & ICCV'23     &   \ding{51}          &    Others             &  1-4          \\ \midrule
ControlVideo~\cite{controlvideo}      & ICLR'24     &  \ding{51}           &  ControlNet~\cite{controlnet}      &    1-5        \\ \midrule
TokenFlow~\cite{tokenflow}      & ICLR'24     & \ding{51}            &    PnP~\cite{pnp}              & 2-1           \\ \midrule
Flatten~\cite{flatten}      & ICLR'24      & \ding{51}            &   Others               & 2-1            \\ \midrule
RAVE~\cite{rave}      & CVPR'24     &  \ding{51}           &    Others              &   1-5         \\ \midrule
Fresco~\cite{fresco}     & CVPR'24     & \ding{51}            &    ControlNet~\cite{controlnet}             & 1-5           \\
\bottomrule
\end{tabular}
}
\caption{Collection of the editing models.}
\label{tab:editing_methods}
\end{table*}

\subsection{Subjective Study}
\begin{figure*}[t]
\centering
\includegraphics[width=1.5\columnwidth]{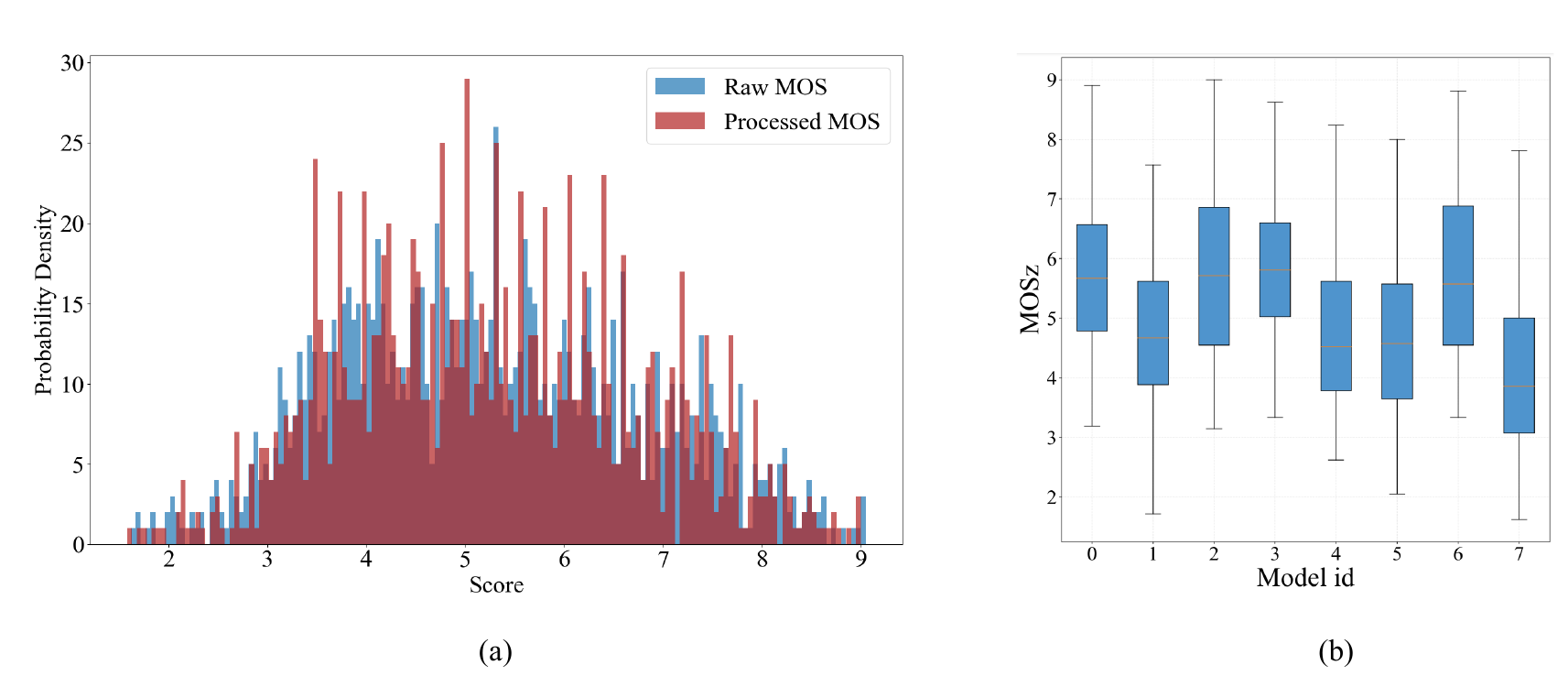} 
\caption{Statistics on MOS. (a) The distribution of the raw/Z-score MOS. (b) Z-score MOS distributions of 8 editing methods.}

\label{fig:mos}
\end{figure*}

According to the ITU standard~\cite{itu}, the number of participants in subjective experiments should be at least 15 to ensure that the results' variance is within a controllable range. For this experiment, a total of 24 human subjects with diverse backgrounds were recruited. During the experiment, the subjects were asked to consider their subjective impressions and evaluate the text-video consistency, source-target fidelity, and quality of the edited videos in a comprehensive manner. The text-video consistency refers to whether the edited content adheres to the given prompt. The source-target fidelity indicates the degree to which the original video and the edited video maintain a certain level of connection. The edited video quality can be assessed from aspects such as temporal and spatial coherence, aesthetics, and technical distortions. When evaluating, all participants rated the videos on a scale from 1 to 10. These participants are all over 18 years old with bachelor's degrees in business, engineering, science, or law. Before started, they receive offline training including cases of varying editing quality beyond the dataset. After that, they rate all samples, taking 5-min breaks every 15 mins to prevent fatigue. 
Following previous works~\cite{dover,t2vqa,stablevideo}, Z-scores are used to normalize the raw MOS values, which could be formulated as
After collecting all raw Mean Opinion Score (MOS) values, we use the Z-score normalization method to eliminate inter-subject differences, which could be formulated as:

\begin{align}
    Z_{m,i} = \frac{X_{m,i} - \mu(X_i)}{\sigma(X_i)},
\end{align}

where $X_{m,i}$ and $Z_{m,i}$ refer to the raw MOS and Z-score of $m$-th video from $i$-th participant, respectively. $\mu(\cdot)$ and $\sigma{(\cdot)}$ represent the mean and standard deviation operators, respectively, and $X_{i}$ is the collection of all MOS from the $i$-th annotator. Then we apply the screening method in BT.500~\cite{bt500} to filter the outliers. The difference between the raw scores and the normalized scores is illustrated in Figure~\ref{fig:mos}.

\subsection{Dataset Analysis}
\begin{figure}[htbp]
\centering
\includegraphics[width=1.0\columnwidth]{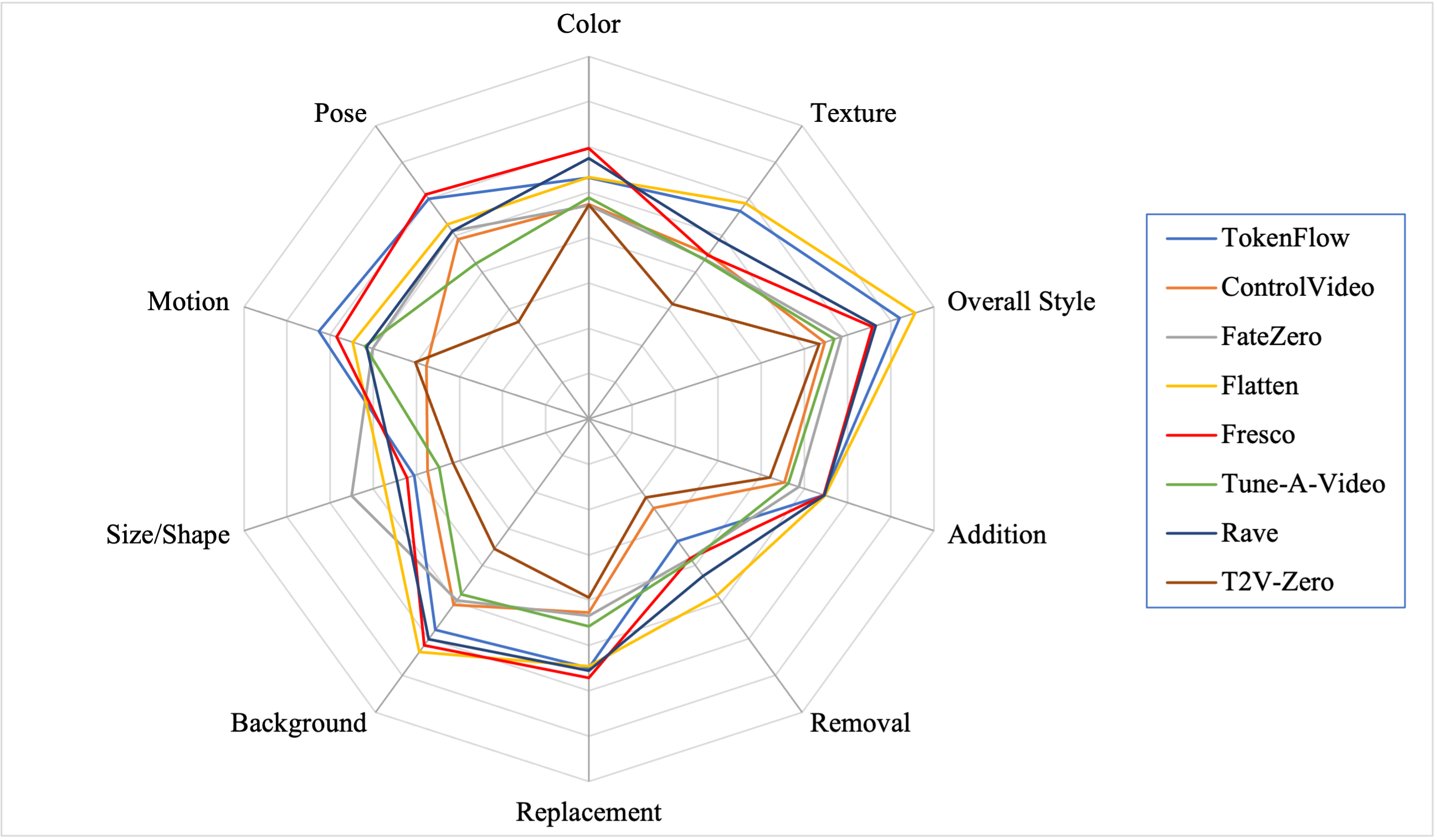} 
\caption{Model performance on different types of prompts.}
\label{fig:editing_comparison}
\end{figure}

We further conducted a detailed analysis of the data in the VE-Bench DB and investigated the scores for the editing results of each model on the collected videos, as shown in Fig.~\ref{fig:editing_comparison}.
We can see that different models have varying capabilities across different types of edits. Currently, models generally score relatively low on ``Removal" tasks, while they excel in stylization instructions. Their proficiency in stylization is likely due to SD (Stable Diffusion) being pre-trained on a large variety of images across different styles, leveraging this prior knowledge. This phenomenon is indirectly validated by some training-free style transfer methods based on SD, such as InstantStyle~\cite{instantstyle,instantstyleplus} and RB-modulation~\cite{rout2024rbmodulation}. Compared to ``Addition", ``Removal" is more challenging, likely because removal tasks involve background reconstruction, which demands stronger semantic understanding and fine-grained feature extraction capabilities. This area might see significant advancements in future editing models.
Additionally, models struggle with shape and size edits, with the exception of FateZero~\cite{fatezero}. This model excels due to its proposed shape-aware Attention Blending technique, which merges the generated shape with the inverted attention of the original image in the latent space, enhancing shape editing capabilities. This technique is one of the core contributions of FateZero, as demonstrated in the original paper, where it achieves superior shape editing performance, supporting the conclusions of this experiment. It is evident that relying solely on native SD models for shape-related edits produces suboptimal results. Future base models could consider incorporating similar shape-aware supervision during training. Furthermore, most open-source video editing models currently use CLIP-based text encoders, which lack the comprehension capabilities of large language models (LLMs). As a result, prompts involving changes in the number of objects often yield subpar outcomes. The motion degree of the source video also impacts editing results—especially for optical flow-based editing methods—since large movements can result in inaccurate optical flow, thereby affecting the final outcome. Complex motion dynamics also pose significant challenges for maintaining frame-to-frame consistency.


\section{VE-Bench QA: Subjective-Aligned Metric for Text-Driven Video Editing}

\begin{figure}[t]
\centering
\includegraphics[width=1.0\columnwidth]{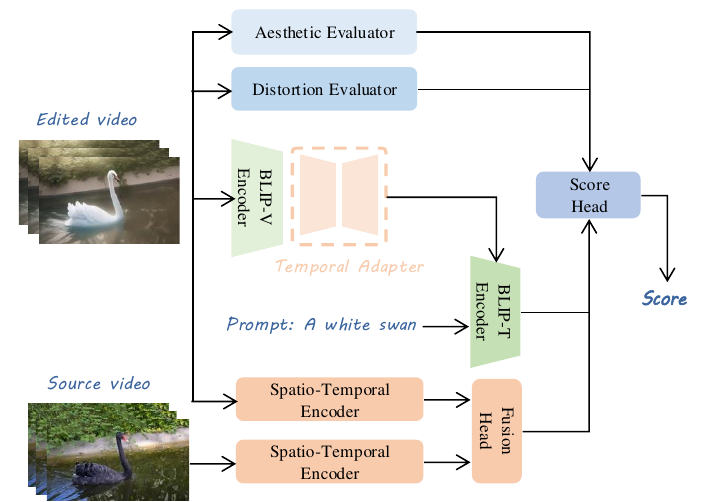} 
\caption{Network architecture of VE-Bench QA.}
\label{fig:VE-Bench-qa}
\end{figure}

Based on the VE-Bench DB, we further developed the VE-Bench Quality Assessment (VE-Bench QA) network, which aligns with human subjective perceptions for evaluating the quality of edited videos, as illustrated in Fig.~\ref{fig:VE-Bench-qa}. The VE-Bench QA evaluates the quality of edited videos from three aspects: (1) Alignment between the edited video and the prompt. (2) Relevance between the edited video and the original video. (3) Quality of the edited video. We will elaborate on each component in the following sections.

\subsection{Video-Text Alignment}
Traditional natural video evaluation methods do not need to consider the alignment between the video and text prompt, which is one reason why they tend to fail when directly applied to AIGC video quality assessment (VQA). Therefore, based on the successful VQA method~\cite{dover}, we incorporate the text branch to model the alignment between the generated content and the corresponding text. Inspired by BLIP~\cite{blip}, we design an effective temporal adapter to extend it to the temporal dimension, as shown in~\ref{fig:VE-Bench-qa}, which could be formulated as,

\begin{align}
    e_{bv} &= \mathbf{F_{bv}}(V^\star), \\
    t_{bv} &= \mathbf{F_{ta}}(e_{bv}), \\
    e_{bt} &= \mathbf{F_{bt}}(p, \mathbf{F_{ca}}(t_{bv})),
\end{align}
where $\mathbf{F_{bv}}$, $\mathbf{F_{bt}}$ refer to the BLIP visual and text encoder, and $\mathbf{F_{ta}}$ represents the temporal adapter. $p$ is the prompt. The derived spatio-temporal feature $t_{bv}$ is then interacted with the text encoder via cross-attention (denoted as $\mathbf{F_{ca}}$).

\subsection{Source-Target Relationship}
Measuring the consistency between the src and dst videos is challenging. There is inherently a connection between the src and dst videos, but there are also significant differences in the pixel space. Therefore, directly using methods like MSE for RGB space measurements has certain limitations.
We design an effective spatiotemporal extractor to project the src-dst videos into a latent space. After concatenating them along the dimension, we obtain a reasonable score estimate through a FFN which could be formulated as,

\begin{align}
    f &= \mathbf{F}(V), \\
    f^\star &= \mathbf{F^\star}(V^\star), \\
    o_s &= \mathbf{H_s}(Concat(f, f^\star)),
\end{align}
where $V$, $V^\star$ denote the original and edited videos, respectively. $o_s$ is the output vector measuring the relevance between source and edited videos. $\mathbf{H_s}$ is the lightweight feed-forward network. $\mathbf{F}$ and  $\mathbf{F^\star}$ denote the spatio-temporal encoder for source and target, respectively. In practice, we test different spatio-temporal backbones and finally choose the Uniformer~\cite{uniformer}. 


\subsection{Visual Quality}
To assess the quality of the edited video, we start from the perspectives used in the previous top-performance method DOVER~\cite{dover}, which evaluates videos based on aesthetics and technical distortion. 
DOVER achieves success in some natural video quality assessment datasets such as~\cite{lsvq, livevqc}.  
In practice, the measurement of aesthetics is implemented via the inflated ConvNext~\cite{convnext} pre-trained on AVA~\cite{ava}, and the distortion is assessed with the Video-Swin~\cite{videoswin} backbone pre-trained with GRPB~\cite{fastvqa}.
At the first stage of training, the backbones from DOVER are frozen and only the parameters of the regression head are updated. In the second stage, all parameters of the visual quality branches (namely, the aesthetic and technical branch) are updated.

\subsection{Supervision}
Following previous works~\cite{dover,fastvqa,trivqa}, we adopt the combination of PLCC (Pearson Linear Correlation Coefficient) loss and rank loss~\cite{rankloss} with the weight of $\alpha$ as the total loss for all branches of the overall network, which could be formulated as follows.
\begin{align}
    L &= L_{plcc} + \alpha \cdot L_{rank},
\end{align}
where $\alpha$ is set to 0.3 in practice.

\begin{figure*}[t]
\centering
\includegraphics[width=2.0\columnwidth]{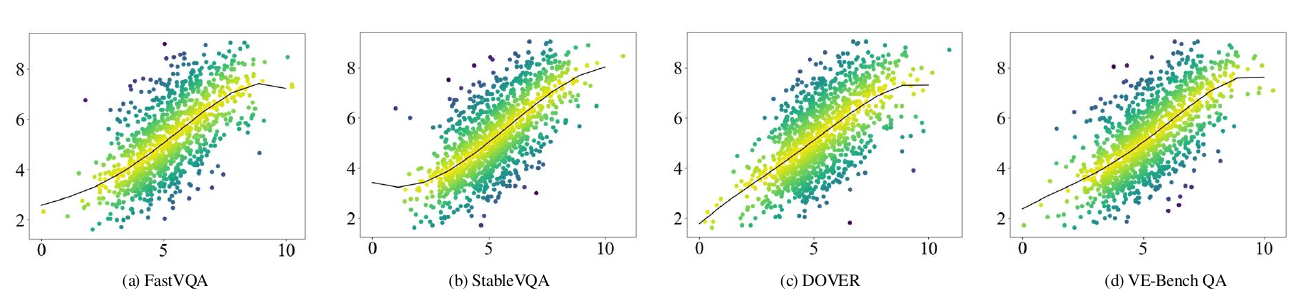} 
\caption{Plots of predicted vs. GT scores. The brightness of scatter points from dark to bright means density from low to high.}
\label{fig:scatter}
\end{figure*}

\begin{table*}[htb]
\centering
\scalebox{1.0}{
\begin{tabular}{lccccc}
\toprule
\multirow{2}{*}{Type} & \multirow{2}{*}{Models} & \multicolumn{4}{c}{VE-Bench DB 10-fold} \\ \cmidrule{3-6}
& & SROCC $\uparrow$ & PLCC $\uparrow$ & KRCC $\uparrow$ & RMSE $\downarrow$ \\ \midrule
\multirow{3}{*}{Zero-shot}
&CLIP-F~\cite{clip} & 0.2284 & 0.1860 & 0.1545 & 4.448 \\
&$S_{edit}$~\cite{flatten} & 0.1686 & 0.1865 & 0.1135  & 3.981 \\
&PickScore~\cite{pickscore} & 0.2266 & 0.2446 & 0.1540 & 1.786 \\
\midrule
\multirow{3}{*}{Fine-tuned}
&FastVQA~\cite{fastvqa} & 0.6333 & 0.6326 & 0.4545 & 1.312 \\
&StableVQA~\cite{stablevqa} & 0.6889 & 0.6783 & 0.4974 & 1.262 \\
&DOVER~\cite{dover}     & 0.6119 & 0.6295 & 0.4354 & 1.311        \\ \midrule
Ours
& VE-Bench QA & \textbf{0.7415} & \textbf{0.7330} & \textbf{0.5414} & \textbf{1.095} \\
\bottomrule
\end{tabular}
}
\caption{Comparison of different methods with VE-Bench QA.}
\label{tab:main_tab}
\end{table*}

\begin{table}[htb]
\centering
\scalebox{0.85}{
\begin{tabular}{llcccc}
\toprule
Experiment    & Method   & SROCC $\uparrow$ & PLCC $\uparrow$ & KRCC $\uparrow$ & RMSE $\downarrow$        \\ \midrule
Baseline & DOVER & 0.6119 & 0.6295 & 0.4354 & 1.311 \\ \midrule
\multirow{2}{*}{Text} 
& CLIP & 0.6379 & 0.6529 & 0.4560 & 1.269 \\
& \underline{BLIP} & \textbf{0.7171} & \textbf{0.7094} & \textbf{0.5193} & \textbf{1.146} \\ \midrule
\multirow{4}{*}{Temporal}
& None & 0.7171 & 0.7094 & 0.5193 & 1.146 \\
& VSwin & 0.7187 & 0.7101 & 0.5197 & 1.143 \\
& MVD & 0.7228 & 0.7143 & 0.5240 & 1.134 \\
& \underline{Uformer} & \textbf{0.7317} & \textbf{0.7252} & \textbf{0.5328} & \textbf{1.116} \\ \midrule
\multirow{3}{*}{Fusion}
& None & 0.7317 & 0.7252 & 0.5328 & 1.116 \\
& MCA  & 0.7330 & 0.7255 & 0.5341 & 1.113 \\
& \underline{Concat} & \textbf{0.7415} & \textbf{0.7330} & \textbf{0.5414} & \textbf{1.095} \\ \midrule
\multirow{2}{*}{Param.}
& w/o & 0.7251 & 0.7174 & 0.5262 & 1.130 \\
& w & \textbf{0.7415} & \textbf{0.7330} & \textbf{0.5414} & \textbf{1.095} \\ \bottomrule
\end{tabular}
}
\caption{Ablation study of the proposed VE-Bench QA.}
\label{tab:ablation}
\end{table}

\section{Experiments}

\subsection{Implementation Details}
We build all models via PyTorch and train them via NVIDIA V100 GPUs. 
Following the 10-fold method~\cite{stablevqa,dover,simplevqa}, all models are trained with the initial learning rate of $1e-3$ and the batch size of 8 on VE-Bench DB for 60 epochs. 
Following DOVER~\cite{dover}, we first fine-tuning the head for 40 epochs with linear probing, and then train all parameters for another 20 epochs.
Adam~\cite{adam} optimizer and a cosine scheduler are applied during training. Following previous works~\cite{dover}, the aesthetic and technical branches of the evaluator are initialized with pre-trained ConvNext~\cite{convnext} and VideoSwin-Tiny~\cite{videoswin} with GRPB~\cite{fastvqa}.

\subsection{Evaluation Metrics}
Following previous works~\cite{dover,t2vqa,stablevqa,simplevqa}, we use four metrics as our evaluation metrics: Spearman’s Rank Order Correlation Coefficient (SROCC), Pearson’s Linear Correlation Coefficient (PLCC), Kendall rank-order correlation coefficient (KRCC), and Root Mean Square Error (RMSE). 

\subsection{Quantitative Results}
We compare our results with advanced evaluation metrics in video editing, including objective metrics~\cite{clip,flatten,pickscore} and state-of-the-art human-aligned Video Quality Assessment (VQA) methods~\cite{stablevqa,fastvqa,dover}. 
The results are shown in Table~\ref{tab:main_tab}.
We further collected a validation set generated by other models~\cite{sliceedit, pix2video}, which are also voted by 24 annotators, and tested on more datasets~\cite{t2vqa}.
Compared with the baseline DOVER~\cite{dover} result, which obtains SRCC and PLCC of 0.4929 and 0.5924. VE-Bench QA attains 0.6008 and 0.6544, respectively. 
We further test models on T2VQA-DB~\cite{t2vqa}, removing the src-dst branch for lack of src videos, as shown in Table~\ref{tab:t2v_results}.
From these, it can be seen that compared to previous traditional VQA methods~\cite{dover, fastvqa,stablevqa} and commonly used objective metrics, VE-Bench QA achieves significantly superior performance in aligning with human subjective perception, surpassing the second by $7.64\%$, $8.06\%$, $8.85\%$, $13.2\%$ in SROCC, PLCC, KLCC, and RMSE, respectively. Compared with our baseline method DOVER, VE-Bench attains higher performance gain, with the improvements of 0.1296,0.1035,0.1060, and 0.216 on SROCC, PLCC, KLCC, and RMSE, respectively. Compared with learning-based methods, the performances of Zero-shot objective measurements are relatively low. Similar situations are quite common in previous works such as~\cite{t2vqa,trivqa}, as the objective quantitative metric struggles to align with human perceptions. Among all Zero-shot methods, PickScore~\cite{pickscore} achieves the top overall performance, which is pre-trained via a reward model with human feedback.
CLIP-F~\cite{clip} achieved comparable SROCC and KRCC metrics to PickScore, but was significantly weaker in terms of PLCC and RMSE metrics. 
$S_{edit}$\cite{flatten}, is obtained by dividing the CLIP-T metric by the Warp-MSE~\cite{pix2video} metric. It reflects the scores on the spatio-temporal quality of the edited video and its alignment with the text prompt to some extent. However, it is not aligned with human subjective perception and obtains limited scores.

\begin{table*}[]
\centering
\begin{tabular}{cccccccccc}
\toprule
\multirow{2}{*}{ Type } & \multirow{2}{*}{ Models }  & \multicolumn{3}{c}{  T2VQA-DB } \\ \cmidrule(lr){3-5}
 && SROCC $\uparrow$ & PLCC $\uparrow$ & KRCC $\uparrow$\\ \midrule
\multirow{5}{*}{ Zero-shot } & CLIPSim~\cite{clip} & 0.1047 & 0.1277 & 0.0702 \\
& BLIP~\cite{blip} & 0.1659 & 0.1860 & 0.1112 \\
& ImageReward~\cite{imagereward} & 0.1875 & 0.2121 & 0.1266  \\
& ViCLIP~\cite{viclip} & 0.1162 & 0.1449 & 0.0781  \\ 
& UMTScore~\cite{liu2023fetv} & 0.0676 & 0.0721 & 0.0453 \\ \midrule
\multirow{5}{*}{ Finetuned } & SimpleVQA~\cite{simplevqa} & 0.6275 & 0.6338 & 0.4466 \\
& BVQA~\cite{bvqa} & 0.7390 & 0.7486 & 0.5487 \\
& FAST-VQA~\cite{fastvqa} & 0.7173 & 0.7295 & 0.5303 \\
& DOVER~\cite{dover} & 0.7609 & 0.7693 & 0.5704 \\
& T2VQA ~\cite{t2vqa} & 0.7965 & 0.8066 & 0.6058  \\ \midrule
Ours 
& VE-Bench QA & \textbf{0.8179} & \textbf{0.8227} & \textbf{0.6370} \\
\bottomrule
\end{tabular}
\caption{Quantitative comparison on T2VQA-DB.}
\label{tab:t2v_results}
\end{table*}


\subsection{Qualitative Results}
We also plot the difference between the predicted scores after training and the MOS scores, as illustrated in Figure~\ref{fig:scatter}.
The curves are obtained by a four-order polynomial nonlinear fitting. 
As the brightness of scatter points grows from low to high, the density goes from low to high. 
From there, it can be intuitively seen that VE-Bench QA has prediction results more aligned with human perception.
We further conducted a qualitative comparison for different score levels in VE-Bench, as illustrated in the supplements, where we present several video examples in VE-Bench DB with varying MOS.

\subsection{Ablation Study}
To further validate the results of each module in VE-Bench QA, we performed detailed ablation experiments on each module, as shown in Table~\ref{tab:ablation}. All results were obtained through 10-fold validation training on the VE-Bench DB with the same experimental hyper-parameter design. The settings we adopted in our final model are underlined.
We first explore different ways of video-text alignment. Here, we experimented with CLIP and fine-tuned the regression head composed of Feed-Forward Networks, which learn the alignment from the cosine similarity of its visual and text Backbone outputs. 
Experiments demonstrate that, although both CLIP and BLIP possess rich vision-language prior knowledge and enhance network performance, BLIP achieved more effective improvements. 
Furthermore, we explored how to effectively model the relevance between the source video and the edited video. 
We attempted to efficiently extract video features and, through experiments conducted on methods such as Video SwinTransformer~\cite{videoswin} (VSwin), Masked Video Distillation~\cite{mvd} (MVD), and Uniformer~\cite{uniformer} (Uformer), we identified suitable feature extractors. 
Additionally, we explored effective ways to fuse features from the source video and the edited video, as presented in Table~\ref{tab:ablation}.
MCA denotes the mutli-head cross-attention, which proves effective in lots of prior works~\cite{meng2022adavit,spcgc,xie2024roi}. We found that concatenation along the dimension is a simple and effective design for the assessment. 
We further ablate the effect of additional parameter, which demonstrates the improvements are not from more parameters.
During these experiments, we could learn that the design of video-text similarity and the focus on the source-target video relationship modeling is of importance to the overall performance.

\section{Conclusion}
In this work, we introduce VE-Bench DB, a subjective-aligned dataset specifically designed for evaluating text-driven video editing, and VE-Bench QA, a novel human-aligned metric for assessing the effects of text-driven video editing. VE-Bench DB features rich video content and detailed editing prompt categories. To the best of our knowledge, VE-Bench DB is the first VQA dataset tailored for text-driven video editing. Furthermore, extensive experiments demonstrate the effectiveness of VE-Bench QA. Compared to traditional metrics commonly used in editing tasks, VE-Bench QA achieves significantly better alignment with human perceptions.

\section{Acknowledgments}
This work was supported by National Science and Technology Major Project (2024ZD01NL00101), Natural Science Foundation of China (62271013, 62031013), Guangdong Provincial Key Laboratory of Ultra High Definition Immersive Media Technology (2024B1212010006), Guangdong Province Pearl River Talent Program (2021QN020708), Guangdong Basic and Applied Basic Research Foundation (2024A1515010155), Shenzhen Science and Technology Program (JCYJ20240813160202004, JCYJ20230807120808017).

\small{
\bibliography{aaai25}
}

\end{document}